\documentclass[conference]{IEEEtran}

\usepackage{cite}
\usepackage{amsmath,amssymb,amsfonts}
\usepackage{algorithmic}
\usepackage{graphicx}
\usepackage{textcomp}
\usepackage{xcolor}
\def\BibTeX{{\rm B\kern-.05em{\sc i\kern-.025em b}\kern-.08em
    T\kern-.1667em\lower.7ex\hbox{E}\kern-.125emX}}
\begin{document}

\title{LLM-based Schema-Guided Extraction and Validation of Missing-Person Intelligence from Heterogeneous Data Sources}

\author{\IEEEauthorblockN{Joshua Castillo}
\IEEEauthorblockA{\textit{Department of Computer Science} \\
\textit{Old Dominion University}\\
Norfolk, Virginia, USA\\
jcast046@odu.edu}
\and
\IEEEauthorblockN{Ravi Mukkamala}
\IEEEauthorblockA{\textit{Department of Computer Science} \\
\textit{Old Dominion University}\\
Norfolk, Virginia, USA\\
rmukkama@odu.edu}
}

\maketitle

\begin{abstract}
Missing-person and child-safety investigations rely on heterogeneous case documents, including structured forms, bulletin-style posters, and narrative web profiles. Variations in layout, terminology, and data quality impede rapid triage, large-scale analysis, and search-planning workflows. This paper introduces the Guardian Parser Pack, an AI-driven parsing and normalization pipeline that transforms multi-source investigative documents into a unified, schema-compliant representation suitable for operational review and downstream spatial modeling. The proposed system integrates (i) multi-engine PDF text extraction with Optical Character Recognition (OCR) fallback, (ii) rule-based source identification with source-specific parsers, (iii) schema-first harmonization and validation, and (iv) an optional Large Language Model (LLM)-assisted extraction pathway incorporating validator-guided repair and shared geocoding services. We present the system architecture, key implementation decisions, and output design, and evaluate performance using both gold-aligned extraction metrics and corpus-level operational indicators. On a manually aligned subset of 75 cases, the LLM-assisted pathway achieved substantially higher extraction quality than the deterministic comparator (F1 = 0.8664 vs. 0.2578), while across 517 parsed records per pathway it also improved aggregate key-field completeness (96.97\% vs. 93.23\%). The deterministic pathway remained much faster (mean runtime 0.03 s/record vs. 3.95 s/record for the LLM pathway). In the evaluated run, all LLM outputs passed initial schema validation, so validator-guided repair functioned as a built-in safeguard rather than a contributor to the observed gains. These results support controlled use of probabilistic AI within a schema-first, auditable pipeline for high-stakes investigative settings.
\end{abstract}

\begin{IEEEkeywords}
Data Harmonization; Heterogeneous Data Sources; Large Language Models; Open-Source Intelligence; Schema Validation; Search and Rescue.
\end{IEEEkeywords}

\section{Introduction}
Time is a critical factor in missing-persons and child-safety investigations. However, the information needed for early decision-making is often distributed across heterogeneous sources (e.g., structured registries, poster-style flyers, narrative Open-Source Intelligence (OSINT) case pages) that differ in formatting, field coverage, and reliability \cite{Bielska2020OSINT, Mider2024OSINT}. Investigators and search coordinators must rapidly consolidate demographic descriptors, last-seen spatiotemporal signals, and narrative circumstances into a coherent “case view” that can support prioritization and search planning.
Research on missing-person analysis increasingly emphasizes data fusion and structured analytics to support evidence-driven decisions \cite{Solaiman2022MissingML, RuizReyes2025MissingReview}. At the same time, practical investigation often begins with unstructured narratives (e.g., posters, witness summaries, police narratives) where key entities must be extracted and normalized before meaningful aggregation is possible \cite{Chau2002PoliceNarratives}. 

In this paper, we describe the Guardian Parser Pack, a schema-guided extraction and validation pipeline, to address this gap. The proposed system transforms heterogeneous case documents into a unified machine-readable representation, allowing for careful downstream analysis, such as hotspot mapping and mobility forecasting \cite{Longley2015GIS, Besenczi2021TrafficMarkov}. It is part of the Guardian system, an end-to-end pipeline that converts raw unstructured case documents into a probabilistic search surface over a wide geographical grid and a set of human-interpretable artifacts: ranked sectors, hotspots, and containment rings for 24/48/72-hour horizons. This paper focuses on document-to-schema extraction, but does not address how the resulting outputs are used downstream in hotspot analysis and mobility forecasting for missing person search planning.

The key research questions that we address in this work are as follows:
\begin{itemize}
\item RQ1 (Extraction \& Normalization): How can heterogeneous missing-person case documents—ranging from structured forms to semi-structured bulletins and unstructured OSINT narratives—be automatically extracted and normalized into a unified schema-driven representation suitable for investigative review and downstream analytics?

\item RQ2 (Validation \& Provenance): To what extent can a schema-first validation and harmonization framework reduce silent extraction errors and preserve provenance to ensure traceability across multi-source inputs?

\item RQ3 (Robust System Design):
How can a hybrid extraction architecture that combines deterministic rule-based parsing with constrained LLM-assisted methods be designed to achieve robustness, auditability, and graceful degradation under document layout variability and temporal drift?

\item RQ4 (Evaluation \& Trade-offs):
What are the practical trade-offs in coverage, completeness, and reliability between rule-based and LLM-assisted extraction approaches, particularly for fields embedded in narrative text versus explicitly labeled content, and how do these trade-offs affect downstream spatial use cases such as geocoding and hotspot analysis?
\end{itemize}



The remainder of the paper is organized as follows. In section 2, we describe related work. Section 3 describes the overall system architecture along with details of data flows and processing in the system. Section 4 contains details of the implementation. Section 5 provides an illustrative example of the process. In section 6, we summarize the results. Section 7 aligns the research requirements (RQ1-RQ4) with the accomplished results of the project. Finally, section 8 provides a summary of our contributions and the proposed future work.
\section{Related work}
Information extraction from investigative text has long targeted the conversion of police narratives into structured entities for analysis \cite{Chau2002PoliceNarratives}. Contemporary NLP toolchains support robust preprocessing and entity recognition, particularly in Python ecosystems 
\cite{Bird2009NLTK}, and industrial Natural Language Processing (NLP) libraries have standardized practical pipelines for tokenization and extraction (Explosion/spaCy). For more broadly unstructured document analysis, recent surveys describe end-to-end pipelines that transform messy multi-format documents into structured outputs for downstream analysis, aligned with the Guardian Parser Pack’s document-to-schema pattern \cite{Mahadevkar2024UnstructuredAI}.

For missing-persons and Search and Rescue (SAR) decision support, previous work covers data fusion and analysis \cite{Solaiman2022MissingML}, geospatial profiling and clustering in missing person contexts \cite{Barone2022LocusOperandi, congram2017grave}, and agent-based or probabilistic mapping approaches for search planning \cite{Hashimoto2022LostPerson}. Mobility modeling methods such as Markov-based propagation over transportation networks provide a mathematically grounded way to forecast movement under uncertainty \cite{Besenczi2021TrafficMarkov}, while GIS foundations clarify how spatial data should be represented and evaluated \cite{Longley2015GIS}. These lines of work motivate the need for reliable, standardized structured inputs before modeling is feasible.

Large language models (LLMs) have shown promise for structured extraction and weak supervision, but multiple studies caution that reliability improves when LLMs are constrained to narrow roles (e.g., assisted extraction or repair) rather than unconstrained end-to-end reasoning \cite{Chen2024LLMAnnotator, Ratner2017Snorkel}. This role-restriction stance is also supported by multi-task evaluation evidence showing that LLMs may perform unevenly across distinct academic-text tasks (e.g., summarization vs. critique/scoring), reinforcing the need to define bounded responsibilities and to avoid over-trusting “judge” behaviors in high-stakes workflows \cite{Li2025LLMEval}.

In weak supervision specifically, LLMs can be used to generate supervision signals that enable scalable text classification without gold labels, aligning with Guardian’s cautious use of probabilistic AI where the corpora of missing-persons labeled are limited \cite{Zeng2022WeakSupervision, Ratner2017Snorkel}. Where retrieval is used (e.g. pulling exemplar fields, canonical formats, or prior similar cases to stabilize extraction prompts), retrieval-augmented prompting is also relevant as a strategy to reduce overreliance on parametric memorization and improve generalization under sparse or noisy input \cite{Chen2023RAPL}.

Ethical frameworks emphasize transparency, accountability, and minimization of harm in sensitive AI applications \cite{Floridi2019AIPrinciples}, and humanitarian guidance highlights risks in applying OSINT and AI to missing-person searches \cite{ICRC2025MissingPeopleTech}. The Guardian Parser Pack adopts these recommendations by enforcing schema constraints, validating outputs, recording provenance, and treating LLM outputs as candidates subject to deterministic checks.


\section{System Architecture}
\begin{figure*}[h]
\includegraphics[width=\textwidth,height=\textheight,keepaspectratio]{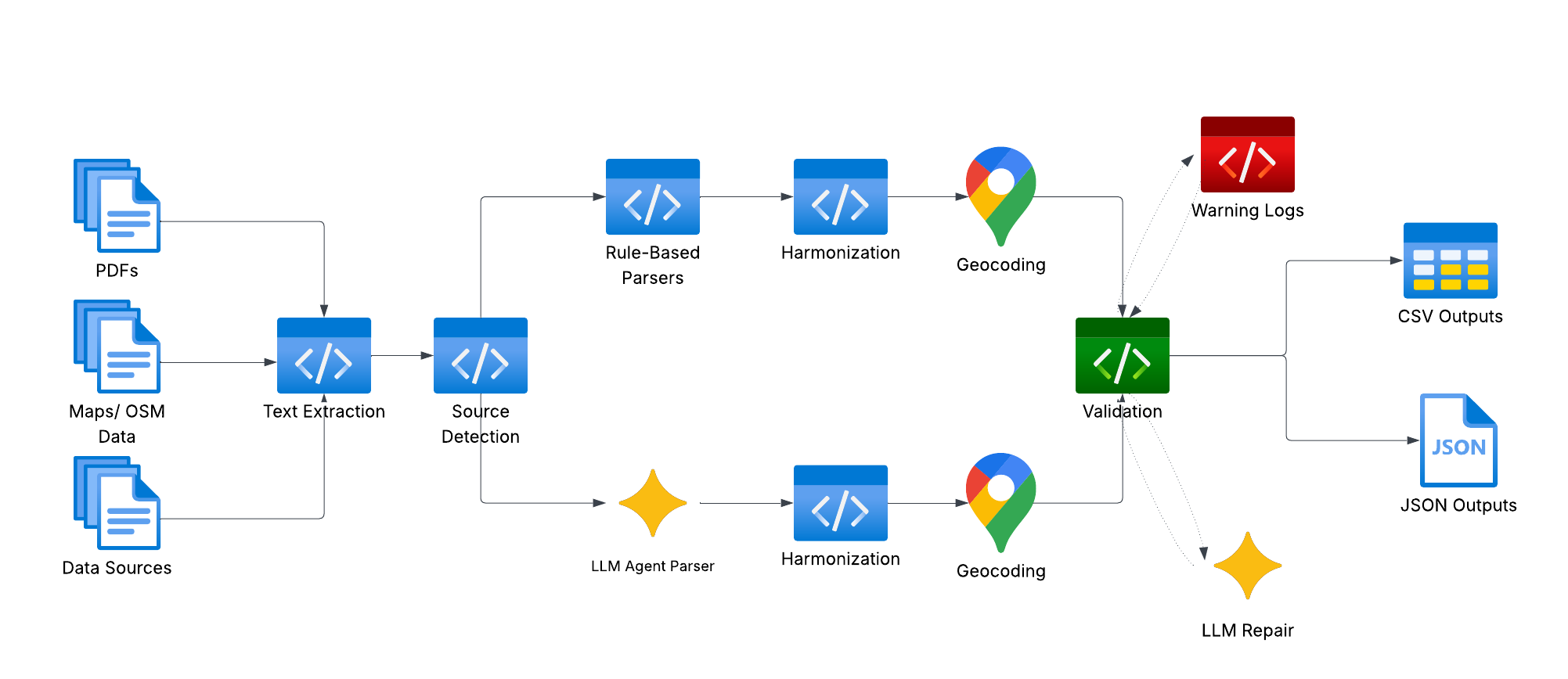}
\caption{Overall System Architecture of the Guardian Parser Pack (Dual-Path Extraction with Shared Services.)}
\label{fig}
\end{figure*}

\begin{figure}
\includegraphics[width=5cm,height=5cm]{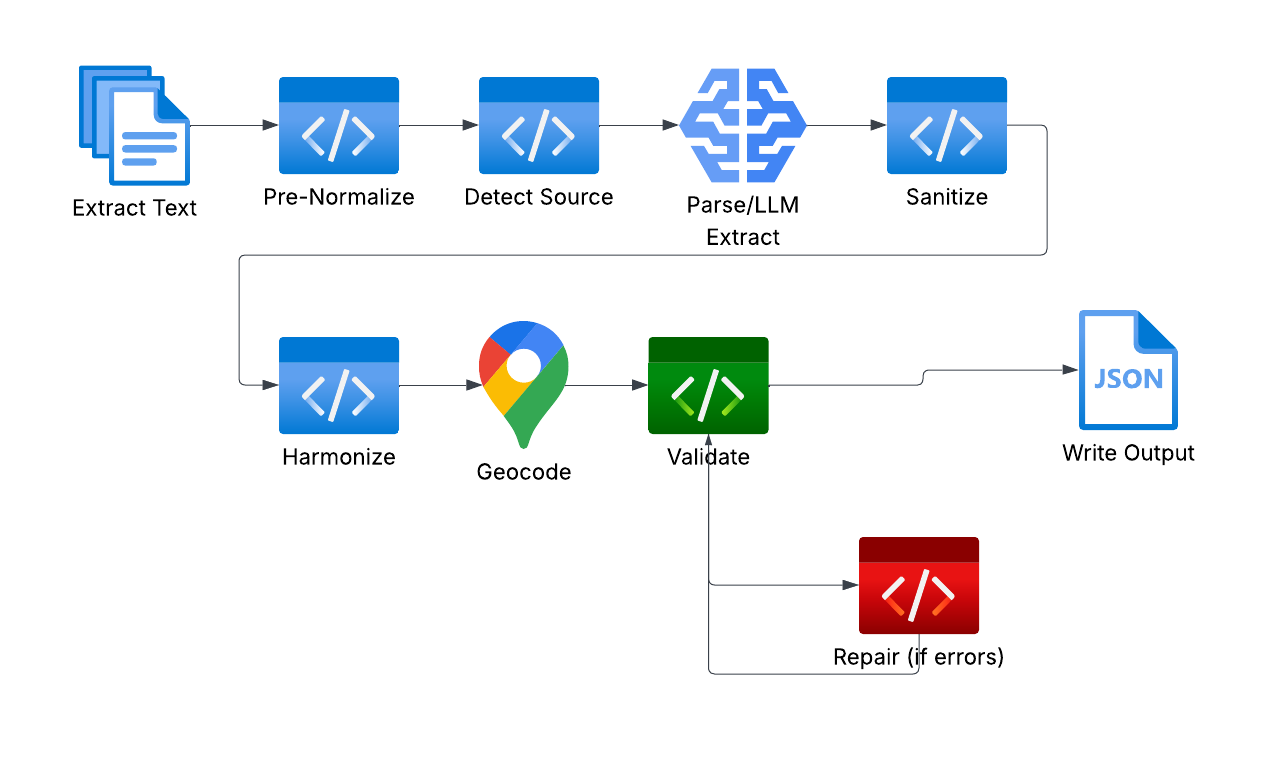}
\caption{Core Parsing and Reasoning Pipeline (Extraction, Harmonization, Validation, and Repair Loop).}
\label{fig}
\end{figure}

Figure 1 presents the overall system architecture of the Guardian Parser Pack system. It is a dual-path document-to-schema pipeline that converts heterogeneous missing-person case PDFs into a unified machine-readable representation. The architecture deliberately separates deterministic extraction (rule-based parsing) from probabilistic extraction (LLM-assisted parsing), while keeping downstream normalization, geocoding, and validation shared across both paths. This separation supports two competing requirements common in investigative settings: stable, reproducible behavior when document layouts are known, and increased coverage when layouts drift or when critical fields are embedded in narrative text. Figure 2 further exemplifies the flow in the system. 
We now describe different functionalities of the system along with the description of the system components involved in carrying out this functionality. 

\subsection {Inputs and Upstream Context (PDFs, Data Sources, Maps/OSM)}

The pipeline begins with heterogeneous case intelligence that primarily arrives as PDF documents, but is produced by a broader set of upstream data sources and may be accompanied by auxiliary maps/OSM resources. In operational missing-person workflows, PDFs are the dominant exchange format for public-facing and inter-agency case information, spanning structured registry exports, poster-style bulletins, and narrative OSINT profiles. 

The “Data Sources” box captures the upstream publishers and repositories that generate these PDFs (e.g., registries, law-enforcement bulletins, and OSINT pages). This matters because source-specific conventions strongly influence both parser routing (which extraction strategy is appropriate) and quality expectations (what fields are likely present, how dates/locations are formatted), motivating explicit source detection and provenance capture rather than treating all inputs as interchangeable text. Finally, “Maps/OSM Data” represents an optional geographic context used to interpret and validate location mentions. This matters because location strings in case documents are often incomplete (“near Route 1,” “left from Norfolk area”), ambiguous (multiple cities with the same name) or jurisdiction-specific, and downstream spatial analytics require consistent grounding in coordinates and administrative regions; map and gazetteer context helps bound plausibility during geocoding and supports later stages such as hotspot modeling and mobility forecasting.

\subsection{Text Extraction}
The “Text Extraction" is implemented as a multi-engine cascade designed to maximize compatibility between PDF types. The system first attempts layout-aware extraction and then falls back to simpler extractors, with OCR invoked when the PDF is image-based. This stage matters because every downstream decision—source detection, parser routing, and field extraction—depends on stable text. Without robust extraction, the system risks failing silently (empty text) or producing brittle artifacts (fragmented lines, missing tokens) that undermine both regex matching and LLM prompting.

Figure 2 explains this stage by making explicit the normalization boundary between raw text and parse-ready text: extraction produces content, but pre-normalization makes that content usable. This distinction is critical for heterogeneous PDFs, where the difference between “some text was extracted” and “the text is structured enough to be parsed reliably” can determine success or failure downstream.
\subsection{Source Detection}
The “Source Detection” (or “Detect Source” in Figure 2) performs the rule-based identification of the document’s origin using markers that are relatively stable over time (e.g., organizational labels, form headers, or recurring boilerplate). 
In a heterogeneous corpus, routing is not a cosmetic decision: it determines which extraction assumptions are permitted (explicit labels versus narrative inference), which fields are expected to be present, and which normalization rules should be applied. Source detection also contributes directly to traceability. By recording the detected source label in provenance metadata, the system preserves an auditable explanation for why a given parser was chosen, which is essential when investigators need to interpret omissions or reconcile conflicting fields across sources.
\subsection{Rule-Based Parsers}
The “Rule-Based Parsers” is the deterministic extraction pathway. Here, the system applies source-specific parsing logic—primarily pattern matching and structured section extraction—to produce an initial record populated with fields such as demographic descriptors, last-known location, and narrative circumstances when explicitly labeled. 

This component also establishes a baseline for comparative evaluation against the LLM pathway. Because both pathways ultimately pass through the same harmonization, geocoding, and validation steps, differences observed in final outputs can be attributed primarily to extraction strategy rather than downstream formatting differences.
\subsection{LLM Agent Parser}
The “LLM Agent Parser” is the probabilistic extraction pathway invoked for documents that are irregular, narrative-heavy, or poorly served by deterministic rules. In Figure 2, this corresponds to “Parse/LLM Extract,” where the system produces an initial structured record from text. The LLM is used here as a schema-guided extractor. The prompt constrains the model to produce JSON aligned with the Guardian schema rather than free-form prose, and the system truncates text when needed to respect context limits.

This component is essential because many investigative PDFs embed key facts in narrative form (e.g., “believed to be en route to Maryland or Delaware,” or temporal cues implied by a report date). Deterministic parsing tends to perform best on explicit labels, whereas schema-guided LLM extraction can recover structured fields from prose when labeling is sparse or inconsistent. However, the architecture explicitly avoids treating LLM output as authoritative: the output is treated as a candidate record that must survive downstream sanitization and validation.

\subsection{Harmonization}
“Harmonization” enables extracted data compatible across the rule-based parser path and the LLM agent parser path. This stage maps source-specific conventions into standardized keys, normalizes units and enumerations, and ensures that required schema sections exist even when inputs are incomplete.
Harmonization matters because downstream analytics depend on consistent semantics. A mobility model or hotspot analysis cannot reliably consume spatial fields if the city/state is stored in different formats, timestamps are not normalized, or demographic attributes vary in naming and typing. By early enforcing a schema-first representation, the system reduces the risk that downstream models will amplify upstream inconsistencies—an especially important consideration in high-stakes investigative settings.
\subsection{Geocoding}
The “Geocoding” (“Geocode” in Figure 2) enriches the schema-aligned record with coordinates derived from 
location text when explicit latitude/longitude are absent. Geocoding performs place-to-coordinate resolution with caching and biasing strategies. 
The shared geocoding service ensures both the rule-based and LLM-derived records receive consistent spatial enrichment under the same assumptions, reducing confounds in evaluation.
\subsection {Validation}
The "Validation” component (“Validate” in Figure 2) acts as the quality gate that distinguishes extraction from reliable structure. Validation checks whether the record is well-formed and plausible. This stage matters because heterogeneous documents often yield partial or noisy extraction results, and without validation the system can produce “valid-looking” outputs that contain type errors, misplaced fields, or structurally inconsistent records that are costly to debug downstream.

The architecture intentionally treats validation differently across the two paths. In the rule-based pathway, validation violations produce warning logs, emphasizing throughput and reproducibility even when some records are incomplete. In the LLM pathway, validation is used as a stricter acceptance criterion: records are expected to pass validation before being written, and failures can trigger repair attempts. This difference reflects the design philosophy that probabilistic extraction requires stronger guardrails to prevent uncontrolled schema drift.
\subsection {Warning Logs}
The “Warning Logs” component externalizes validation outcomes and extraction anomalies. Warnings provide structured error reporting that supports audit and debugging. Warning logs matter because they preserve visibility into failure modes without forcing the pipeline to halt. In investigative workflows, partial information can still be operationally useful, but only if its limitations are explicit. Logging also supports iterative parser maintenance under document drift, since recurring warnings can identify which sources or fields are degrading over time.
\subsection{LLM Repair}
``LLM Repair” is a conditional step tied to validation failures on the LLM pathway and corresponds to “Repair (if errors)” in Figure 2. Repair performs validation-guided correction (constrained regeneration conditioned on errors). Rather than re-extracting the entire record, the system supplies the validator’s error messages alongside the current JSON and instructs the LLM to make minimal edits needed to satisfy schema requirements.
This stage matters because it operationalizes a controlled use of LLMs: the model is not asked to “improve the case” or infer missing facts, but to correct structural and typing violations so the record becomes consumable safely downstream. It also reduces manual cleanup costs by automatically resolving common failures (e.g., invalid types, missing required fields) while keeping the repair bounded by explicit validator feedback. In the present evaluation, repair was not triggered because all LLM-pathway outputs passed initial schema validation. Accordingly, the repair loop should be understood as a bounded fallback mechanism for schema-invalid outputs rather than as a completeness-recovery mechanism or a source of the gains reported in Section VI.
\subsection{CSV and JSON Outputs}
CSV supports quick human review. CSV matters because investigative and research workflows often require rapid scanning, filtering, and manual inspection. A flattened view allows reviewers to audit extraction quality at scale without needing specialized tooling to browse nested JSON.

JSON provides nested schema-aligned serialization suitable for downstream parsing and modeling. This output matters because downstream components (e.g., spatial clustering, mobility forecasting, or provenance-aware analytics) benefit from hierarchical organization and typed fields, which are difficult to preserve faithfully in a flat table.
\subsection{Provenance-Preserving Transformation}
The dataflow pipeline moves from ``PDF Source” to ``Extracted Text,” then to ``Extracted Fields,” and finally to a ``Schema Aligned Record” that is emitted both as a JSONL record and as a flattened CSV row. The dotted provenance line emphasizes that traceability is maintained throughout the transformation: the system is designed to preserve information about where fields came from (source, location in the document, and timestamps when available), enabling auditors to reconstruct how a final value was produced.

\section{Implementation}
\subsection{Inputs, Data Sources, and Preprocessing Assumptions}
The system ingests PDF documents drawn from heterogeneous sources commonly encountered in missing-person contexts: structured registry exports, poster-style bulletins, and narrative OSINT case profiles. In our development corpus, representative sources included NamUs and NCMEC records, Virginia State Police (VSP) bulletins, FBI posters, and narrative profiles from The Charley Project (National Missing and Unidentified Persons System, n.d.; National Center for Missing \& Exploited Children, n.d.; Virginia State Police, n.d.; Federal Bureau of Investigation, n.d.; The Charley Project, n.d.). 

More broadly, our sources selection and documentation practices align with OSINT collection guidelines that emphasize tool transparency and source evaluation \cite{Bielska2020OSINT, Mider2024OSINT}. Inputs are assumed to be well-formed PDF files, but may be text-based or image-based scans. Text extraction therefore uses a fallback cascade: (i) a layout-aware PDF text extractor, (ii) a secondary extractor when layout extraction fails, and (iii) OCR for image-based PDFs.

After extraction, text is pre-normalized to reduce variance caused by PDF formatting. The pre-normalization step collapses repeated whitespace, standardizes line endings, and removes common control characters. For documents containing multiple cases (e.g., bulletin-style multi-entry PDFs), the pipeline can split the document into case segments before parsing.

\subsection{Programming Language(s) and Core Libraries}
The Guardian Parser Pack is implemented primarily in Python. Python is used for PDF handling, regex-based extraction, schema validation, and geospatial utilities; this choice aligns with standard NLP and data processing practices \cite{Bird2009NLTK} and supports integration with GIS tooling when needed \cite{Longley2015GIS}. Geospatial transformations and spatially indexed inspection workflows are supported by common Python geospatial tooling, where appropriate. The optional LLM-assisted path uses a backend abstraction layer that can connect to local or hosted inference endpoints while still returning schema-targeted JSON. When retrieval is used to stabilize prompt formats or enforce source-specific conventions, retrieval-augmented prompting principles further motivate design as a mitigation against overreliance on memorized patterns \cite{Chen2023RAPL}. We have employed Gemini-2.5-flash for the  LLM parser and Gemini-2.5-pro for the LLM Repair.

\subsection{Output Structure and Format}
The system outputs two synchronized artifacts: (1) JSONL, one schema-aligned JSON object per case record, suitable for downstream parsing, indexing, and modeling; and (2) CSV, a flattened representation where nested keys are expanded into dot-separated columns to facilitate human review and quick filtering.

The core schema sections include demographic; spatial; temporal; narrative\_osint; outcome; and provenance. The record is keyed with a case identifier and divided into schema sections. The spatial block contains both a human-readable location and coordinates; coordinates may be derived from geocoding when the source does not provide explicit lat/lon. The temporal block uses International Organization for Standardization (ISO) 8601 timestamps with an explicit timezone field indicating the source context. The narrative summary is intended for triage; it is not treated as ground truth and is maintained alongside provenance metadata indicating the extraction path used. Downstream systems can filter, cluster, or map cases based on validated spatial fields only after schema checks confirm numeric types and plausible ranges.
\section{Illustrative Example}
To illustrate the end-to-end behavior of the Guardian Parser Pack, we trace a NamUs PDF (MP102335) from raw document (Figures 3 and 4) to a schema-aligned case record (Figures 5 and 6). The input is a form-style report containing labeled demographic attributes (e.g., name, sex, age range, height/weight, race/ethnicity), temporal cues (“Date of Last Contact,” “NamUs Case Created”), spatial context (“Culpeper, Virginia 22701,” “Culpeper County”), and narrative blocks describing circumstances, clothing and other distinctive features.

\begin{figure}
\includegraphics[width=10cm, height=8cm]{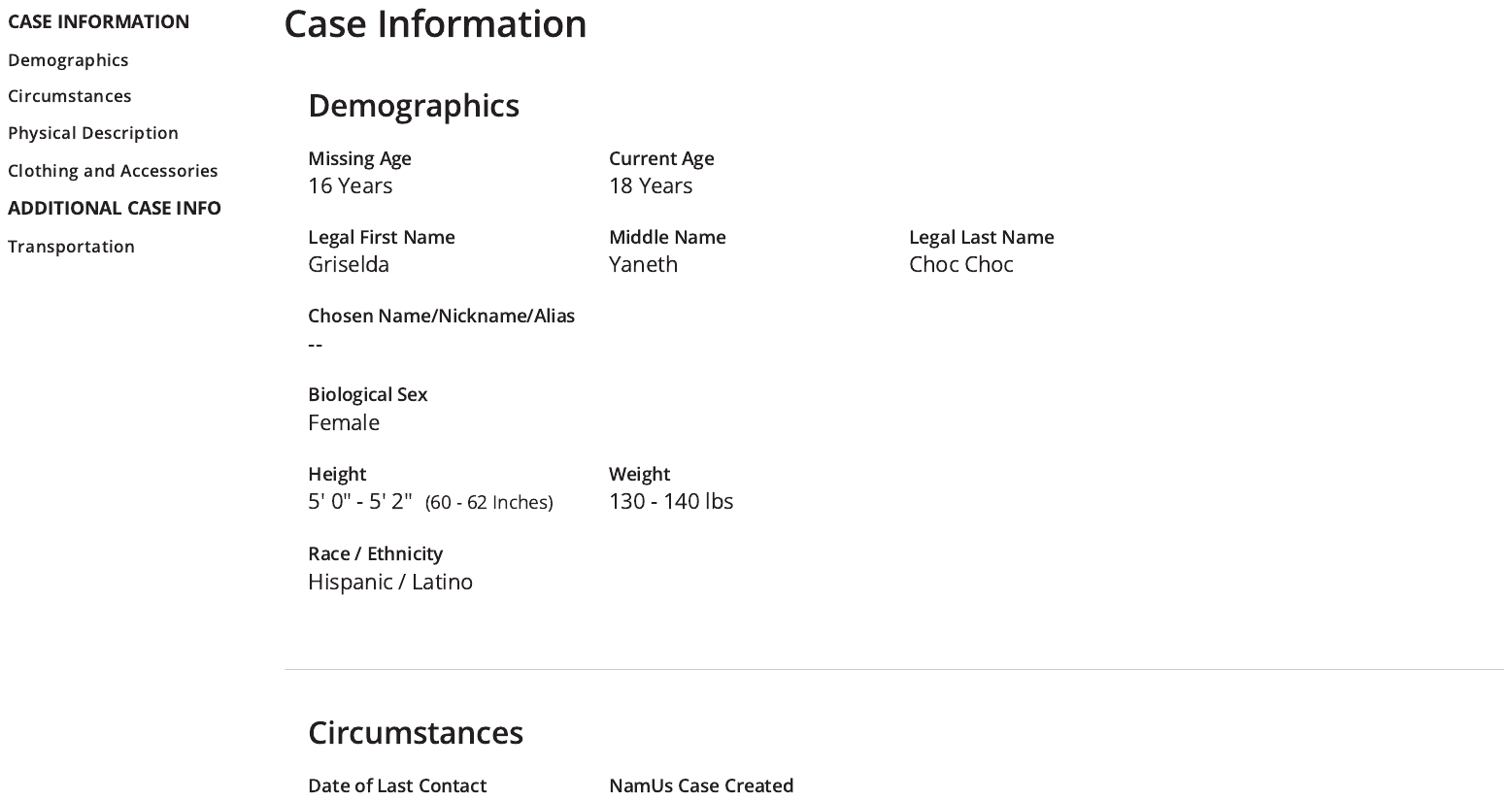}
\caption{Example Missing Person Document-Page 1}
\label{fig}
\end{figure}
\begin{figure}
\includegraphics[width=10cm, height=8cm]{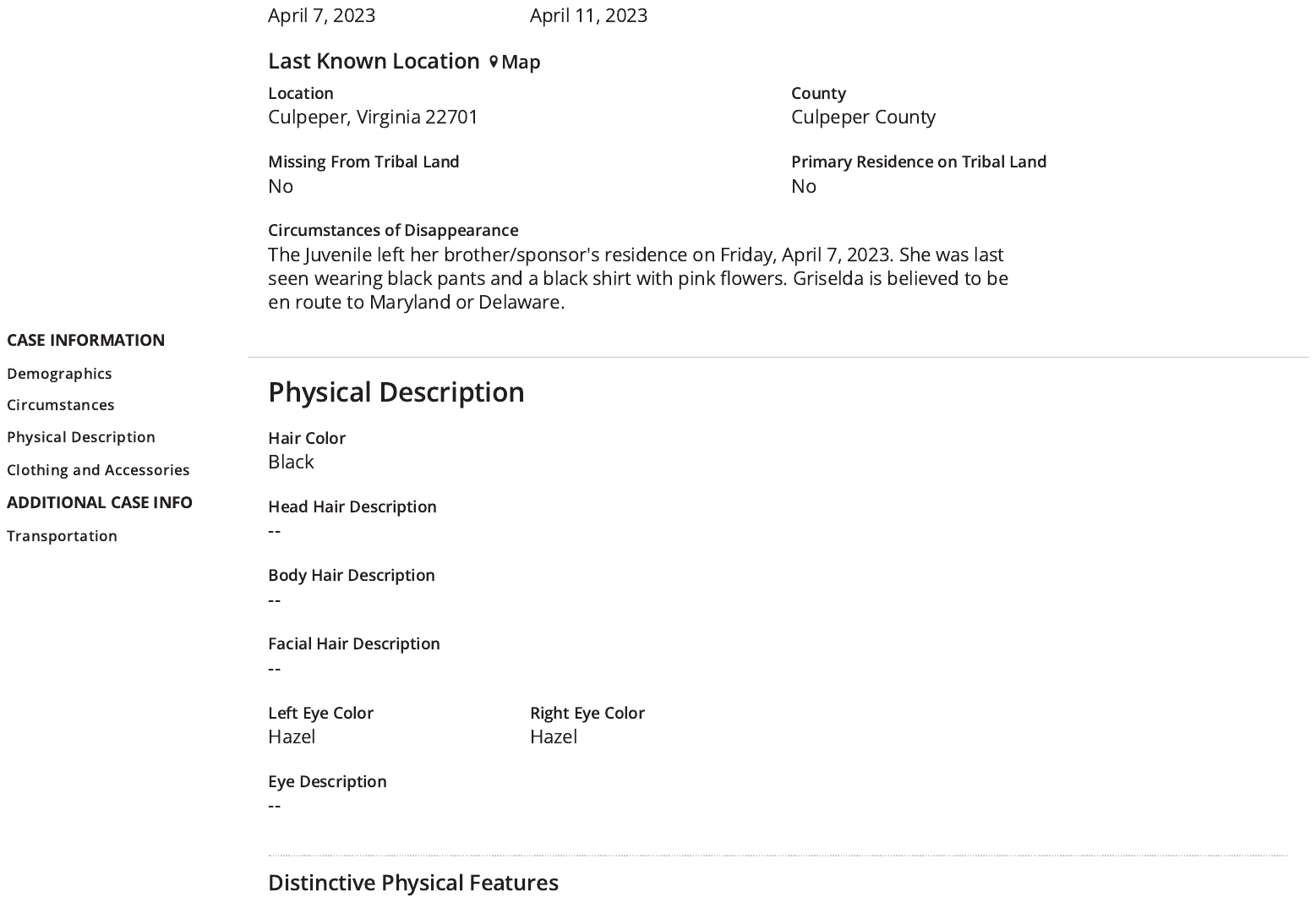}
\caption{Example Missing Person Document-Page 2}
\label{fig}
\end{figure}
\begin{figure}
\includegraphics[width=8cm, height=10cm]{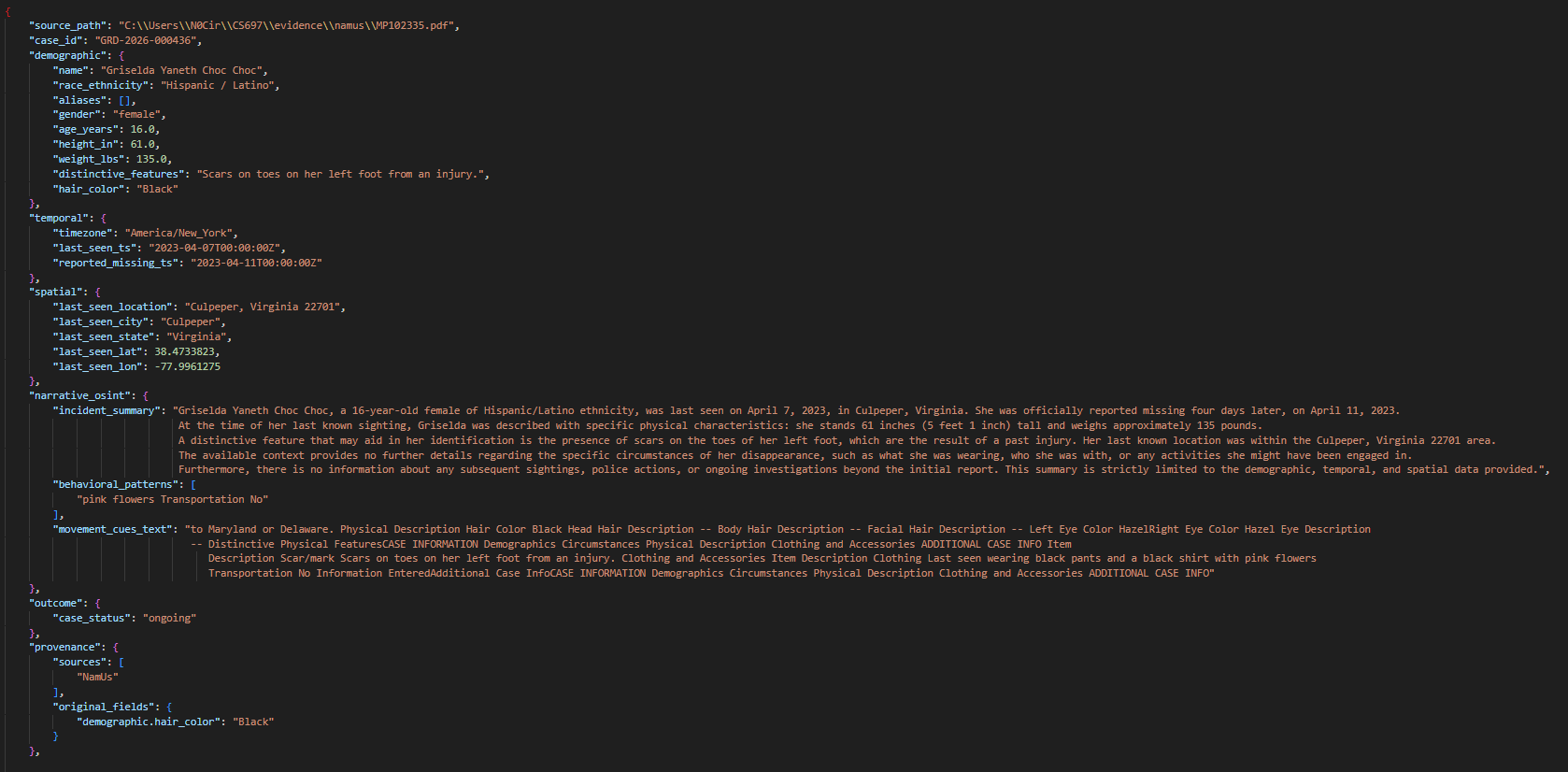}
\caption{LLM Parser Path Output for the Example Document}
\label{fig}
\end{figure}

\begin{figure}
\includegraphics[width=8cm, height=10cm]{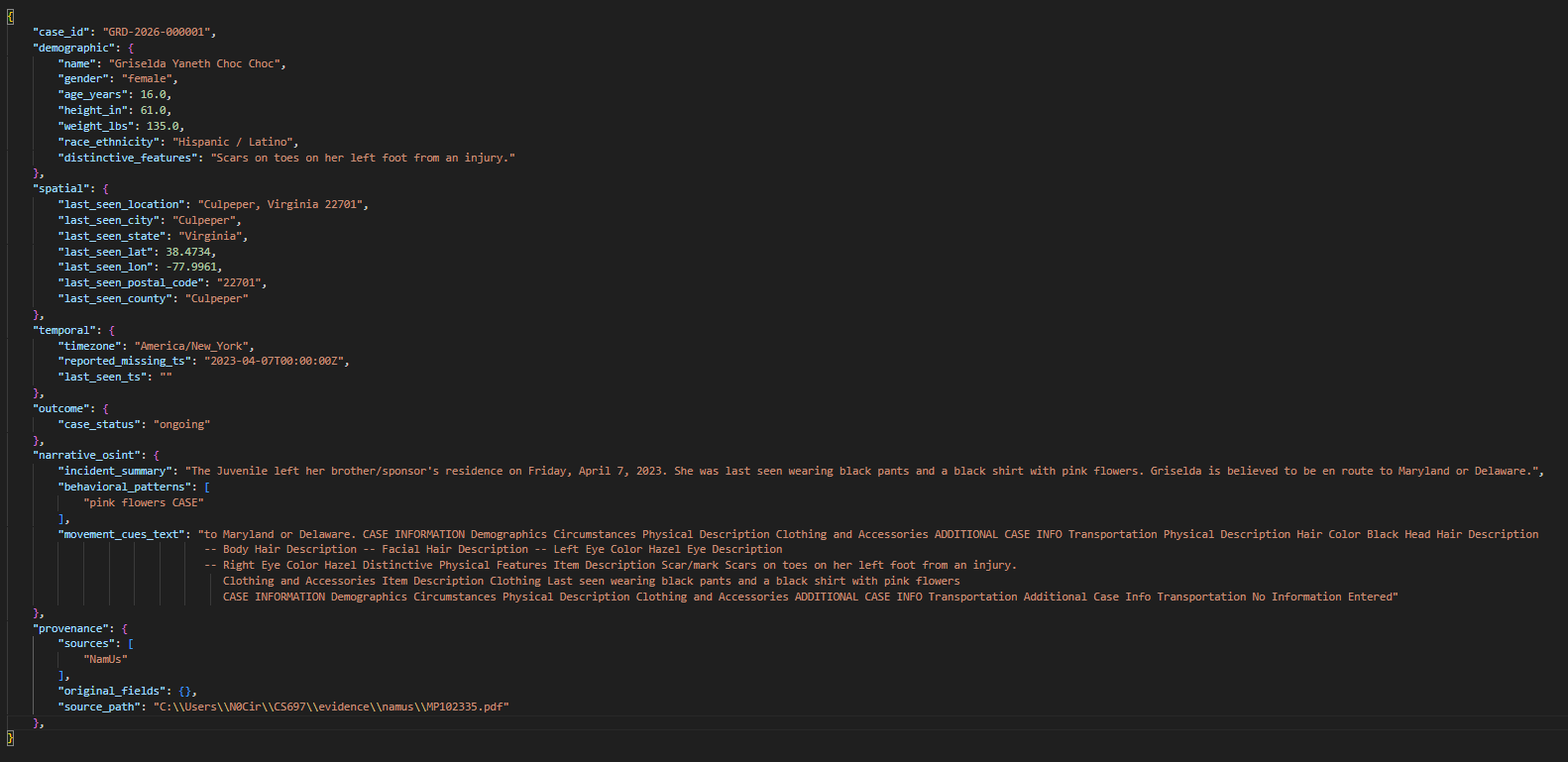}
\caption{Rule-based Parser Path  Output for the Example Document}
\label{fig}
\end{figure}

The pipeline first converts the input PDF document (Figure 4) into machine-readable text (text extraction), using a multi-engine fallback so that native PDF text or OCR-derived text can be processed without changing downstream logic. 

Next, the system determines which source family produced the document (source detection) by matching distinctive markers; in this example, NamUs-specific headings and field labels allow deterministic routing. 

Once routed, the system extracts key fields into a structured draft record (parsing), either by deterministic label/regex matching for form-like NamUs layouts or by constrained model-based extraction when layouts are irregular (rule-based parsing or LLM-assisted extraction). In this case, the parser lifts the core identifiers and descriptors—subject name, female sex, age, height/weight ranges, race/ethnicity, last-known location, county and the narrative ``Circumstances of Disappearance,” into their corresponding schema sections, ensuring that narrative text remains separated from typed attributes.

The draft record is then standardized so that the same concepts are represented in the same way across all sources (sanitization and harmonization). 
If the document provides only a human-readable place string, the system fills in missing coordinates from that string (geocoding). This matters because spatial modeling and mapping require numeric latitude/longitude, so “Culpeper, Virginia 22701” can be converted into coordinates with caching to improve repeatability and reduce external lookups (geocoding with cache).

Before any record is accepted, the system checks that it satisfies the required structure and constraints (schema validation) and, when the LLM path is used, can attempt minimal corrections guided by validator errors (validator-guided repair). 

Finally, the validated case is written in two synchronized formats (final structured outputs): JSONL preserves the nested schema (e.g., demographic/spatial/temporal/narrative/provenance blocks) for programmatic use, while CSV flattens the same content for rapid human review and filtering (newline-delimited JSON and flattened tabular export). 
Figure 5 shows the JSON output of the system from the LLM parser path connecting to the input document in Figure 4. Figure 6 shows the same for the Rule-based parser path.
\section{Results}
The results should be interpreted with caution, as missing-person documents vary widely across sources and over time, and ground-truth labels for “correct extraction” are often unavailable. To address this, we report both gold-aligned extraction metrics on a manually aligned subset and corpus-level operational indicators over a larger batch of parsed records.

\subsection{Qualitative Outcomes}
We list the following distinctive outcomes from this project towards the rescue and search process in missing person investigations.
\begin{itemize}
\item Robustness to heterogeneous layouts: the multi-engine extraction cascade improves practical ingestion by handling both text-based PDFs and scanned documents; The OCR fallback reduces total failure in image-based inputs \cite{Mahadevkar2024UnstructuredAI}.
\item Auditability through schema-first design: records produced by either path are harmonized and validated against the same schema, reducing silent inconsistencies—a key safety property when downstream models can amplify upstream noise \cite{Floridi2019AIPrinciples, ICRC2025MissingPeopleTech}.
\item Conservative LLM integration: LLM-assisted extraction is used as a candidate generator constrained by sanitization and validator-guided repair, consistent with recommendations to bound LLM roles in high-stakes document workflows \cite{Chen2024LLMAnnotator} and with evidence that LLM capability varies by task framing, motivating narrow role assignment and evaluation \cite{Li2025LLMEval}.
\end{itemize}

\subsection {Quantitative Extraction Evaluation}
To provide a direct extraction-quality comparison, we evaluated both pathways on a gold-aligned subset of 75 cases. The deterministic comparator used the rule-based pathway under the same downstream harmonization, geocoding, and validation pipeline as the LLM-assisted pathway. On this subset, the LLM-assisted pathway achieved precision = 0.8664, recall = 0.8664, and F1 = 0.8664, while the deterministic comparator achieved precision = 0.2593, recall = 0.2641, and F1 = 0.2578. For structured field accuracy, the LLM-assisted pathway achieved 0.8810 compared with 0.3090 for the deterministic comparator. These results indicate that the LLM pathway substantially improves extraction quality on heterogeneous and narrative-heavy inputs. The system tracks measurable quality-control signals: schema pass rate after harmonization (and repair when enabled), field completeness for key attributes (e.g., name, last\_seen\_ts), geocoding success rate (non-zero, plausible coordinates), and repair frequency for LLM outputs. These indicators support iterative improvements and align with GIS evaluation practices that consider spatial plausibility in addition to textual correctness \cite{Longley2015GIS, RuizReyes2025MissingReview}.

\subsection{Repair Effectiveness and Validation Outcomes}
The system tracks schema-validation outcomes separately from extraction completeness. In the current evaluation, all LLM-pathway outputs passed schema validation before repair, yielding a pre-pass rate of 100.00\%, a post-pass rate of 100.00\%, and a repair rate of 0.00\%. This is consistent with the design of the repair loop: repair is triggered only by schema/type/format violations, not by schema-valid incompleteness. Thus, in this run, validator-guided repair functioned as a safeguard rather than as a contributor to the reported performance gains.

\subsection{Field Completeness and Geocoding Indicators}
On a representative batch containing 517 parsed records from each pathway, the LLM-assisted pathway achieved higher aggregate key-field completeness than the deterministic pathway (96.97\% vs. 93.23\%). Geocoding success was 100.00\% for both pathways, and plausible geocoding rates were 95.36\% for the LLM pathway and 94.78\% for the deterministic pathway. Because both pathways share the same geocoding and harmonization services, these differences are modest compared with the larger extraction-quality differences observed in identity and narrative-derived fields.

\subsection{Performance Considerations}
Runtime behavior differed substantially between pathways. The deterministic pathway achieved a mean runtime of 0.03 s per record, while the LLM-assisted pathway required 3.95 s per record on average. This confirms the expected trade-off: the rule-based path is more suitable for high-throughput batch ingestion when layouts are stable, whereas the LLM-assisted path is more appropriate for narrative-heavy or irregular documents where extraction quality and field recovery are more important than latency.

\section {Alignment with Research Questions}
As mentioned in the introduction, the project was intended to address four research questions (RQ1-RQ4). Having described the system and its implementation, we now discuss the outcomes in terms of these questions.
\begin{itemize}
\item RQ1 (Extraction \& Normalization): The proposed system directly addresses schema-driven extraction and normalization by combining robust text acquisition (multi-engine extraction with OCR fallback), source-aware routing (rule-based source detection), and a shared harmonization layer that maps heterogeneous outputs into a canonical Guardian schema. The dual-path design ensures that the system can extract from both explicitly labeled forms (via deterministic parsers) and narrative-heavy documents (via schema-guided LLM extraction), while still producing a unified representation suitable for investigative review and downstream analytics. 
\item RQ2 (Validation \& Provenance): The architecture emphasizes schema-first validation as a guardrail against silent extraction errors, using JSON Schema checks to enforce structure, types, and required sections across both pathways. This reduces the likelihood that malformed records propagate unnoticed into downstream modeling and supports an explicit representation of missingness by requiring fields to be present (often with safe defaults or empty values) rather than implicitly absent. The provenance is preserved through source labeling and source-path tracking.
\item RQ3 (Robust System Design): The robustness of the system is derived from layered fallback strategies and the limited use of probabilistic components. Text extraction gracefully degrades from layout-aware parsing to simpler extraction and finally to OCR, reducing total ingestion failure across diverse PDFs. When source layouts are stable, deterministic parsers provide reproducible behavior; when layouts drift or narratives dominate, the LLM pathway increases coverage, but remains bounded by sanitization and validation gates. 
\item RQ4 (Evaluation \& Trade-offs): The shared harmonization, geocoding, and validation stages make the two pathways directly comparable. Quantitatively, on a gold-aligned subset of 75 cases, the LLM-assisted pathway substantially outperformed the deterministic comparator in precision, recall, and F1 (0.8664 vs. 0.2578 F1). On 517 parsed records per pathway, the LLM approach also improved aggregate key-field completeness (96.97\% vs. 93.23\%), while geocoding success remained high in both pathways. The principal cost of this gain was latency: the deterministic pathway averaged 0.03 s per record, whereas the LLM pathway averaged 3.95 s per record. These results show that the rule-based path is preferable for stable, explicitly labeled layouts and high-throughput use, while the LLM-assisted path is preferable when critical information is embedded in narrative text. 
\end{itemize}
\section{Summary and Future Work}
The Guardian Parser Pack provides a schema-guided pipeline for converting heterogeneous missing-person documents into structured records suitable for investigative triage and downstream geospatial modeling. Its key design choice is a dual-path extractor with shared deterministic validation: rule-based parsing provides reproducibility and speed for known layouts, while LLM assistance improves extraction quality and completeness for narrative-heavy or irregular documents, constrained by sanitization and schema validation, with validator-guided repair retained as a fallback for schema-invalid outputs.

Limitations include document drift that can break source-specific parsers, uncertainty introduced by geocoding from partial location text, and residual risk of LLM over-generalization. These risks are mitigated by schema validation, provenance preservation, and conservative use of LLM outputs as candidates rather than authoritative facts \cite{Floridi2019AIPrinciples, ICRC2025MissingPeopleTech}. The decision to restrict LLM components to bounded roles is further supported by multi-task evaluations showing that LLM performance is not uniform across task categories, especially when asked to behave as evaluators or critics \cite{Li2025LLMEval}.

Future work includes stronger uncertainty quantification for geocoding and narrative-derived fields, benchmark datasets and evaluation protocols for extraction in disappearance analysis \cite{RuizReyes2025MissingReview}, controlled weak-supervision pipelines to create labeled data \cite{Ratner2017Snorkel}, and tighter integration with geospatial SAR modeling methods that propagate uncertainty through mobility and hotspot forecasting \cite{Hashimoto2022LostPerson, Besenczi2021TrafficMarkov}. Where retrieval is expanded (e.g., using similar historical cases to stabilize extraction and normalization), retrieval-augmented prompting offers a principled approach to reduce brittle “memorization” behaviors and improve generalization \cite{Chen2023RAPL}. A current limitation is that repair is schema-validation driven rather than completeness-driven; schema-valid but incomplete records are therefore not automatically repaired, which remains a direction for future work.

\bibliographystyle{IEEEtran}
\bibliography{main}
\end{document}